\documentclass[nohyperref]{article}
\usepackage{microtype}
\usepackage{graphicx}
\usepackage{subfigure}
\usepackage{booktabs} 
\usepackage{hyperref}
\usepackage{oz}
\usepackage{dblfloatfix}
\usepackage{enumitem}
\setlist[itemize]{leftmargin=*}
\newcommand{\R}{\mathbb{R}}

\usepackage[accepted]{icml2023}

\newcommand{\denotation}[1]{\llbracket #1\rrbracket}

\newcommand{\abs}[1]{\lvert#1\rvert}

\usepackage{savesym}
\savesymbol{implies}
\savesymbol{proof}
\usepackage{amsmath}
\usepackage{mathtools}

\usepackage{amsthm}

\newcommand{\expect}{\ensuremath{\mathop{\mathbb{E}}}}

\newcommand*{\img}[1]{%
    \raisebox{-.3\baselineskip}{%
        \includegraphics[
        height=\baselineskip,
        width=\baselineskip,
        keepaspectratio,
        ]{#1}%
    }%
}

\DeclareMathOperator*{\argmin}{arg\,min}
\DeclareMathOperator*{\argmax}{arg\,max}
\makeatletter
\newcommand*{\rom}[1]{\expandafter\@slowromancap\romannumeral #1@}

\usepackage[capitalize,noabbrev]{cleveref}

\theoremstyle{plain}

\theoremstyle{definition}

\theoremstyle{remark}

\usepackage{stmaryrd}

\icmltitlerunning{ROAP}

\begin{document}

\twocolumn[
\icmltitle{From Perception to Programs:
Regularize, Overparameterize, and Amortize}

\icmlsetsymbol{equal}{*}

\begin{icmlauthorlist}
\icmlauthor{Hao Tang}{cornell}
\icmlauthor{Kevin Ellis}{cornell}
\end{icmlauthorlist}

\icmlaffiliation{cornell}{
Cornell University}

\icmlcorrespondingauthor{Hao Tang}{haotang@cs.cornell.edu}

\vskip 0.3in
]

\printAffiliationsAndNotice{} 

\begin{abstract}
	We develop techniques for synthesizing neurosymbolic programs.
 Such programs mix discrete symbolic processing with continuous  neural computation.
 We relax this mixed discrete/continuous problem and jointly learn all modules with gradient descent,  and also incorporate amortized inference, overparameterization, and a differentiable strategy for penalizing lengthy programs. Collectedly this toolbox improves the stability of gradient-guided program search, and suggests ways of learning both how to parse continuous input into discrete abstractions, and how to process those abstractions via symbolic code.
\end{abstract}

\section{Introduction}
	We seek steps toward AI systems that learn to symbolically process perceptual input.
	Consider, for example, a system which learns to infer the 3D structure of objects: starting from pixels, it must infer low-level  symbols (curves, parts), and then organize them according to symbolic relationships (symmetry, part repetitions, part hierarchy).
	Or, consider a system which learns to control a moving object that navigates around obstacles: starting from sensory data (lidar, RGBD), it must first parse the world (into objects, proximities, freespace), and then compute trajectories using high-level computations (PID controllers, etc.).
	Similar perceptual-symbolic problems arise when learning structured world models from pixels, inferring instructions from natural language, or constructing visual analogies.
	We propose framing such tasks as \textbf{neurosymbolic program synthesis}: learning neural components that extract symbols from perception, and synthesizing programs to further process those symbols with more complex computations.

 Our ultimate goal is to develop general methods that could, we hope, apply to challenging neurosymbolic tasks like those previously mentioned.
	We take the stance that symbols should be grounded in perception, and that symbol processing should be implemented by learnable program-like representations.
	However, we also propose that rather than hand-code a preordained set of primitive symbols, AI systems should learn to carve the perceptual world into their own discretization.
	What constitutes a `symbol' may vary across domain and across datasets, and can be hard for human engineers to anticipate. By jointly learning the symbols, as well as synthesizing the programs that operate on them, we hope to side-step the pitfalls associated with hand-engineered representations.

 Delivering on the above promises requires synthesizing neurosymbolic programs, which poses unique technical challenges.
 Unlike conventional programs, which are discrete, a neurosymbolic program has both continuous weights and discrete program structure, both of which must be synthesized.
 In addition to solving a mixed discrete/continuous problem, synthesizing a neurosymbolic program is severely underconstrained.
 It is under constrained because it is not clear what parts of the problem should be handled by symbolic processing, and what should be handled by neural networks.
 Because neural nets are universal function approximators, they can in theory satisfy any program learning problem, at least on the training data.

 Our main technical contribution is a suite of methods for circumventing the above two challenges.
 We assume a multitask setup where the learner is exposed to a variety of neurosymbolic programming tasks.
 Having multiple tasks  introduces extra constraints, and also allows learning across tasks how to search for programs.
 Hence multitasking can address both the ill-posed nature of the problem, and also the intractable search aspect due to the mixed discrete/continuous nature of the problem.

Concretely, our method trains a neural search policy to synthesize neurosymbolic programs.
It uses a differentiable interpreter to backprop gradients from the desired program output all the way back to the parameters of the search policy.
We overparametrize the program search space to ease continuous optimization, but this overparametrization leads to bloated programs with too much code, hence we regularize the length of the programs to produce concise, interpretable code.
We therefore call our method ROAP (Regularize, Overparametrize, Amortize, for Programs).
\begin{figure*}[h]
    \centering
    \includegraphics[width=\textwidth]{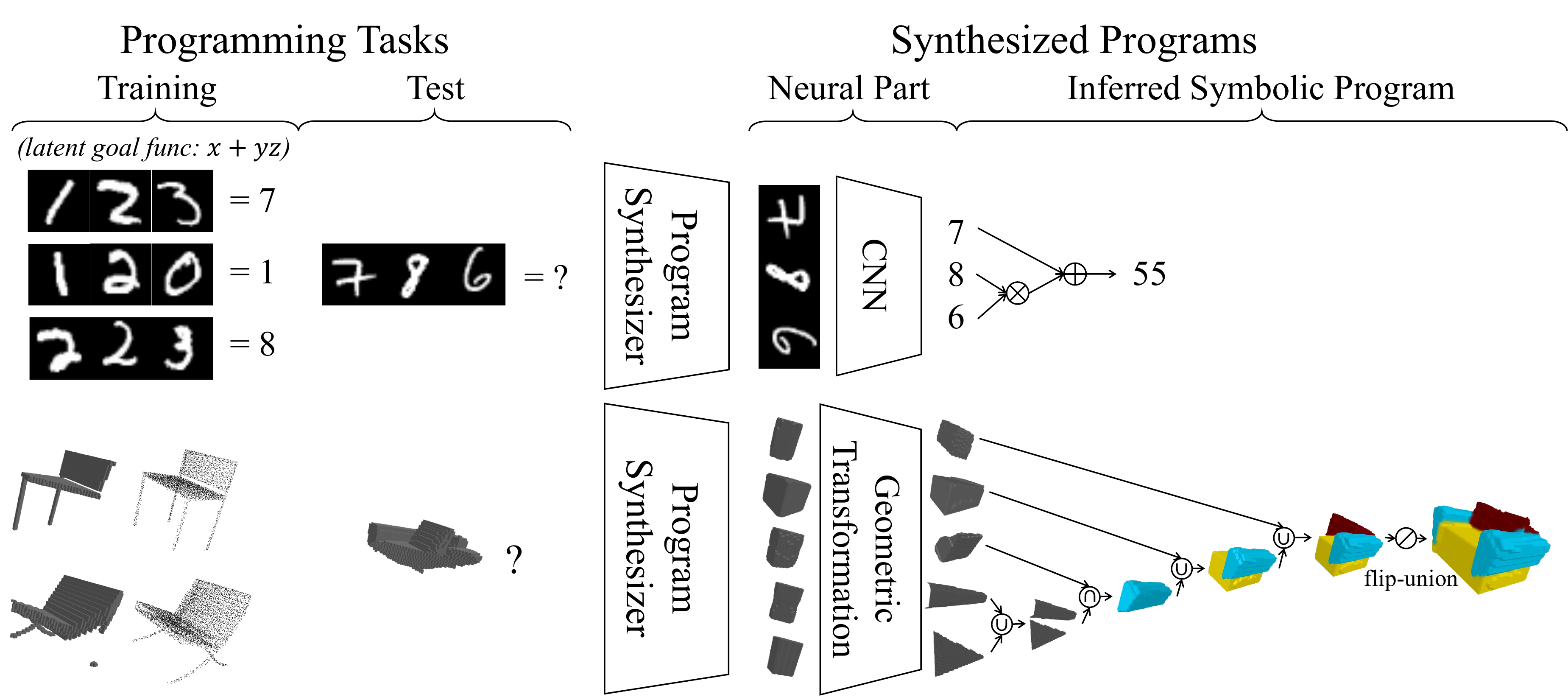}
    \caption{\textbf{Top, CIFAR-MATH domain.} System synthesizes symbolic equations, but instead of learning the equations from concrete numbers, it inputs CIFAR-10 images, where each image category (dog, cat, frog, ...) has been mapped to a digit from 0--9. Note: we show MNIST above for ease of understanding.
    \textbf{Bottom, graphics domain.}
    Given a partial observation of a 3D shape, system learns to infer a 3D graphics program completing the shape. Note: For CIFAR-MATH we show one task. For 3D we show several tasks.}
    \label{fig:setting}
\end{figure*}

We apply ROAP to two different domains (Fig.~\ref{fig:setting}).
Our CIFAR-MATH domain is a harder version of a classic proving ground for neural logic programming~\cite{manhaeve2018deepproblog}, modified to include program synthesis.
On it, we show that ROAP can synthesize arithmetic equations while at the same time learning to parse images into symbolic digits.
Our 3D-Reconstruction domain involves synthesizing graphics programs that algebraically transform and combine  neural geometric primitives, and can be used to decompose 3D shapes and infer missing geometry.
In total our work makes the following contributions:
\begin{itemize}
\item A synthesis method for neurosymbolic programs.
ROAP works without supervising on source code, and so doesn't require a training set of programs, unlike e.g. Codex~\cite{DBLP:journals/corr/abs-2107-03374}.
This is important because it allows unsupervised learning over large datasets, even if those datasets are not designed for program synthesis.
ROAP also does not require pretraining the `neural' part of a neurosymbolic program, unlike~\cite{10.1145/3453483.3454047}.
ROAP \emph{does} require a Domain Specific Language, which restricts the space of programs and imparts human prior knowledge.
\item Comparison against four prior neurosymbolic synthesis methods~\cite{shah2020learning,gaunt2016terpret, valkov2018houdini, cui2021differentiable}.
To the best of our knowledge, the  neurosymbolic program synthesis field lacks  direct comparisons among these prototypical methods.
\end{itemize}

\section{Problem statement \& Technical background}
\textbf{Definitions: Architecture, parameters, denotation.} A neurosymbolic program has both a \textbf{symbolic program architecture} $\mathbf{\alpha}$, and also \textbf{continuous parameters} $\mathbf{\theta}$.
Each architecture comes from a set $\mathcal{A}$ of possible architectures.
We can instantiate a fixed architecture with different continuous parameters, and we write $\alpha_\theta$ for the program with  architecture $\alpha$ and parameters $\theta$.
We assume a \textbf{denotation operator} $\denotation{\cdot }$, which takes a program $\alpha_\theta$ and outputs what the program executes to. Generally, $\denotation{\alpha_\theta}$ is a function.

An example of synthesizing a neurosymbolic program  is optimizing for the architecture $\alpha\in \mathcal{A}$ and parameters $\theta\in \R^d$ minimizing a loss function over training data $\mathcal{D}$: 
\begin{align}
\alpha, \theta&=\argmin_{\substack{\theta\in \R^d\\\alpha\in \mathcal{A}}} \sum_{(x,y)\in \mathcal{D}} \text{Loss}
\big( y,\denotation{\alpha_\theta}\left( x \right) \big)
\end{align}
This is challenging because it involves optimizing over discrete $\alpha$ (from combinatorially large $\mathcal{A}$) and continuous  $\theta$ (which is potentially high dimensional).
The trick of \textbf{relaxation} is to convert this mixed discrete-continuous problem into a purely continuous one, and then optimize with continuous methods.
Intuitively, relaxations index the space of architectures using continuous weights that interpolate between discrete structures:

\textbf{Definition: Relaxation.} Architectures $\mathcal{A}$ and denotation $\denotation{\cdot }$ admit a  \textbf{$k$-dimensional relaxation} when the architectures are represented as $k$-dimensional vectors ($\mathcal{A}\subset\R^k$) and we can take the denotation of any such k-dimensional vector, even ones not in $\mathcal{A}$, which means $\R^k\subseteq\text{domain}(\denotation{\cdot })$.
 
 There are many relaxation approaches differing on what exactly the denotation means as an embedding `interpolates' between architectures.
Some relaxations define an approximate probabilistic semantics and interpret the $k$-dimensional vector as a vector of probabilities~\cite{si2019synthesizing,gaunt2016terpret,chaudhuri2010smooth}.
Others use schemes reminiscent of fuzzy logic~\cite{evans2018learning}, or form linear combinations of discrete subprograms~\cite{sahoo2018learning}.
Either way, solving the relaxation typically proceeds by finding a continuous vector using gradient-based optimization, and then discretizing that vector to the closest symbolic architecture.

\textbf{Amortized inference: Learning to search.} 
The idea behind amortized inference~\cite{gershman2014amortized} is to learn to search for programs (``infer'' programs). 
Instead of directly optimizing over the space of programs, amortized inference in this context means optimizing a policy that probabilistically generates programs, conditioned on a particular programming task to solve. 
Typically the policy is trained across many tasks so that it learns to generate programs that solve each task~\cite{devlin2017robustfill}.

\section{Method}
We assume a training corpus of neurosymbolic programming tasks, $\mathcal{T}$.
Each such task $t\in \mathcal{T}$ is specified by a dataset $\mathcal{D}_t$ of input-output pairs, $(x, y)$.
For example, in the CIFAR-MATH domain, each input $x$ is a triple of CIFAR-10 images, while each output $y$ is a real number, and the task $t$ is a collection of (image-triple, scalar) input-outputs.
In the 3D reconstruction domain, 
$x$ is a point in 3D space and $y$ is either 1 or 0, depending on if $x$ is inside or outside the object, respectively;
meanwhile $t$ is voxels, possibly with missing or noisy data.

\textbf{Amortization \& Parameter Sharing.} We start with an objective function for amortized inference that optimizes the parameters of a policy, $\phi$, to increase the probability of generating a program architecture that has low loss.
We also optimize the parameters $\theta$ of the neural networks invoked by these symbolic program architectures, ultimately  minimizing $\mathcal{L}(\theta, \phi)$ shown 
below:
\begin{align}
\mathcal{L}(\theta, \phi)=\expect_{\substack{t\sim \mathcal{T}\\ \alpha \sim \pi_\phi(\cdot |t)}}\left[ \sum_{(x,y)\in \mathcal{D}_t}\text{Loss}(y, \denotation{\alpha_\theta}(x)) \right] \label{eq:objective}
\end{align}
Already, this framing helps address one issue with synthesizing neurosymbolic programs:
Each program can invoke learned neural networks, but only ones using \emph{shared} parameters $\theta$.
Thus having multiple tasks introduces extra constraints on $\theta$, preventing the system from solving everything with  monolithic neural networks.
The alternative possibility of optimizing task-specific continuous parameters ($\theta_t, \forall t\in \mathcal{T}$) would not introduce these extra constraints.

\textbf{Gradient estimation via Relaxation.} 
Learning to search for programs requires optimizing the search policy parameters $\phi$.
We implement our policies as neural networks, so we are interested in taking gradients of $\mathcal{L}$ with respect to $\phi$ (and also $\theta$).
While one could use a reinforcement-learning style approach, getting useful training signal from such methods would require serendipitously finding good program architectures from a randomly initialized policy.
Absent pretraining, symbolic programs are hard to randomly guess correctly.
How then can we get training off the ground?

At a high level, ROAP relaxes the symbolic program space; assumes that sampling a program architecture is equivalent to sampling an array of one-hot vectors from categorical distributions; and then finally uses Gumbel-Softmax to backprop through these categorical draws.
In low-level detail, we assume the relaxed program semantics allow backpropagating through the denotation operator.
However, we still have to pass gradients backward through the random sampling from the policy (expectation over $\alpha\sim \pi_\phi(\cdot |t)$ in Eq.~\ref{eq:objective}).
To do this, we assume each symbolic architecture $\alpha$ is encoded as 
$C$ one-hot vectors,\footnote{To get intuition on why this assumption is reasonable, imagine that $\alpha$ is the contents of the input tape of a Universal Turing Machine with tape alphabet $\Sigma$; the UTM's output is $\alpha$'s denotation. If $C$ is the maximum program length, then we need $C$ one-hot vectors of dimension $|\Sigma|$ to encode $\alpha$, because $\alpha$ would be represented by a length $C$ string.}
notated $\left\{ 
\alpha_c \right\}_{c=1}^C$, and the policy $\pi_\phi$ samples an architecture $\alpha$ by drawing from $C$ categorical distributions with parameters  $\{ p^c_\phi(t)\}_{c=1}^C$:
\begin{align}
\pi_\phi(\alpha|t)&=\product_{1\leq c\leq C} \text{Cat}\left( \alpha_c ; p^c_\phi(t) \right)
\end{align}
This licenses rewriting the objective in Eq.~\ref{eq:objective} as
\begin{align}
\expect_{\substack{t\sim \mathcal{T}}}
\expect_{\substack{\alpha_c\sim \text{Cat}\left( \cdot ; p^c_\phi(t) \right)\\\forall 1\leq c\leq C}} \left[ \sum_{(x,y)\in \mathcal{D}_t}\text{Loss}(y, \denotation{\alpha_\theta}(x)) \right]\label{eq:objective2}
\end{align}
At this point we can deploy the well-known Gumbel-Softmax trick~\cite{jang2016categorical}, which offers a low-variance approximation to the above expectation.
Gumbel-Softmax perturbs the raw probabilities  $\{ p^c_\phi(t) \}_{c=1}^C$ with Gumbel-distributed noise, then takes a softmax with a temperature that aneals toward 0.
At 0 temperature,  Gumbel-Softmax exactly implements Eq.~\ref{eq:objective2}.
When the temperature is positive, Gumbel-Softmax produces program architectures $\alpha$ whose constituent ``one-hot'' vectors actually contain multiple positive components.
This causes the relaxed denotation operator to interpolate the behavior of nearby program architectures, yielding stable gradient estimation.

\textbf{(Over)parameterizing the program space.}
We now specify what program architectures look like, and how we parameterize them in terms of one-hot vectors.
We model each program architecture as straightline code:
A sequence of $L$ lines of code, each of which introduces a new variable in scope by applying a function to variables introduced by preceding lines of code (Fig.~\ref{fig:parametrization}).
Each function comes from a Domain Specific Language, which contains components customized to the kinds of programs we expect to synthesize.
Toggling which vector component of $\alpha$ is a 1 corresponds to toggling which function each line of code uses, and which preceding lines are passed as arguments to that function.

To compute the denotation of $\alpha$, given its vectorized encoding, we use a simple dynamic program that memoizes the computation of the value computed by each line of code.
This runs in time quadratic w.r.t. the total lines of code.

Although this parametrization  works reasonably well, there are many alternatives.
We also tried encoding a syntax tree instead of a list of lines of code, but this worked worse.
In general, the classic program synthesis literature is filled with different techniques for `sketching' a large set of possible programs, and then indexing that set with  boolean decision variables~\cite{solar2008program, jha2010oracle}.

\begin{figure}
\begin{tabular}{l}
Expression: $(x+x)\times y$\\
Domain Specific Language:\\
$\qquad f_+(a, b)=a+b\qquad f_\times(a, b)=a\times b$\\ 
Straight Line Code:\\
$\qquad\ell_1\gets x$\\
$\qquad\ell_2\gets y$\\
$\qquad\ell_3\gets f_+(\ell_1, \ell_1)$\\
$\qquad\ell_4\gets f_\times(\ell_3, \ell_2)$\\
Architecture parametrization:\\
$\quad$\begin{tabular}{rcccccccc}
&\multicolumn{2}{c}{\underline{function}}&
\multicolumn{3}{c}{\underline{left arg}}&
\multicolumn{3}{c}{\underline{right arg}}
\\&$f_+$&$f_\times$&$\ell_1$&$\ell_2$&$\ell_3$&$\ell_1$&$\ell_2$&$\ell_3$\\\cline{2-9}
Line 3&\multicolumn{1}{|c}{(1,}&0)&(1,&0)&&(1,&0)&\multicolumn{1}{c|}{}\\
Line 4&\multicolumn{1}{|c}{(0,}&1)&(0,&0,&1)&(0,&1,&\multicolumn{1}{c|}{0)}\\\cline{2-9}
\end{tabular}

\end{tabular}
\caption{Symbolic expressions are built from operators in a Domain Specific Language and represented as straightline code.
Each line of code is parametrized by three one-hot binary vectors specifying a function from the Domain Specific Language, and left/right arguments from earlier lines. 
(The first lines of code simply load variables into scope.)
 The bottommost box shows the 6 one-hot vectors encoding the example expression ($\alpha$ in the paper).}\label{fig:parametrization}
\end{figure}

We now \textbf{overparametrize} the problem by expanding the maximum possible lines of code far beyond what the system needs to solve its programming problems.
This  dramatically increases the dimensionality of $\alpha$, and empirically we found that this significantly improved the convergence properties of  gradient descent when optimizing Eq.~\ref{eq:objective2}.
Without overparametrization, the system is prone to falling into poor local minima. 
We speculate that overparametrization helps for our problem for similar reasons as to why it helps for deep networks: (1) that it is harder to get trapped in higher dimensional spaces, because there is likely at least one direction which leads to lower loss, and (2) with more parameters there is a higher chance of a randomly initialized subnetwork falling within the basin of a good optimum, known as the lottery ticket hypothesis~\cite{frankle2018lottery}.

\begin{figure}
    \centering
    \includegraphics[width=\columnwidth]{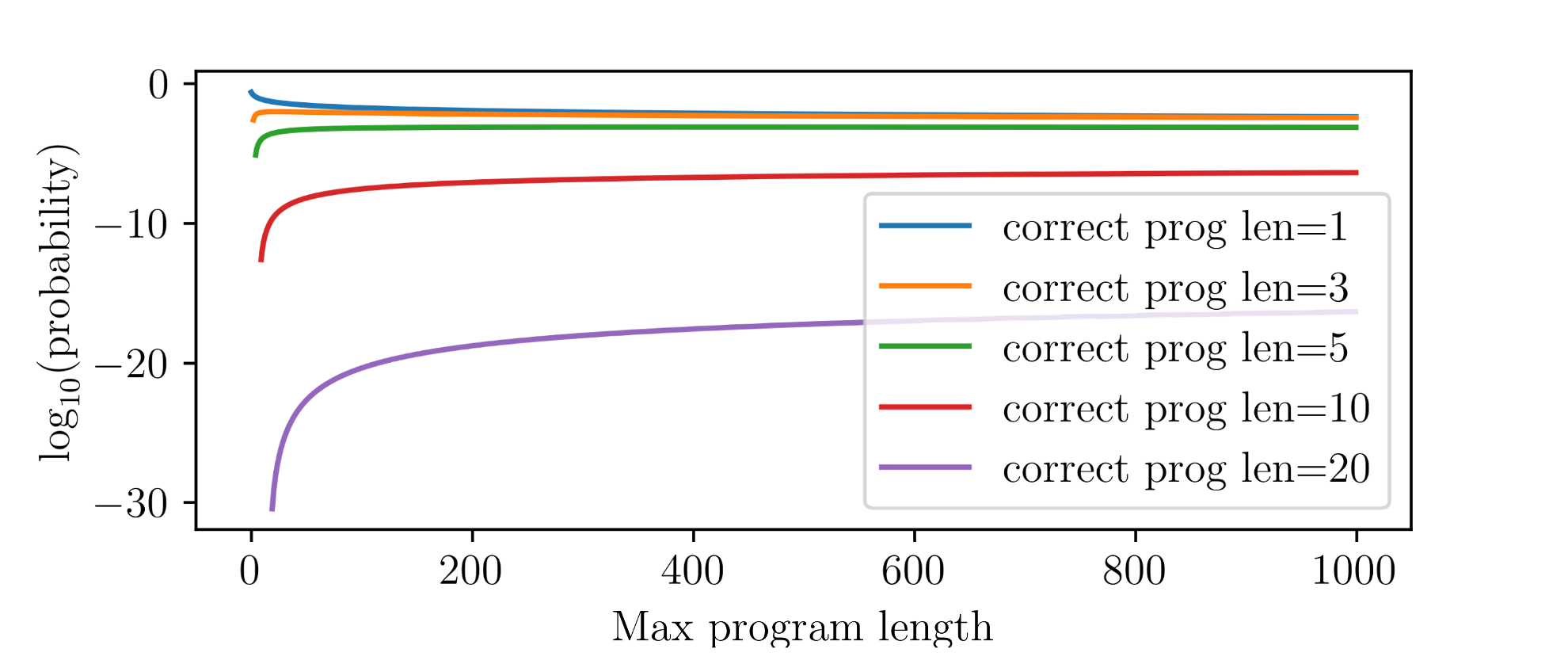}
    \caption{Simulation results showing probability of randomly initializing to a correct program, while varying max program length and  target program length}
    \label{fig:theory}
\end{figure}

Speculatively, if lottery-ticket type behavior accounts for the success of overparametrization in our setting, then we might expect that increasing the maximum lines beyond the needed sizes actually \emph{increases} the probability that randomly initialized weights encode the correct program.
We built a simplified theoretical model of randomly initialized program architectures (Appendix Sec.~\ref{sec:appendixtheory}).
Using this model we calculated the probability of a random network containing the correct program (Fig.~\ref{fig:theory}).
Across a range of different ground-truth program lengths, this probability saturates around a few tens of lines of code.
In agreement with this analysis, we empirically found $L=30$ max lines of code worked well on both of our domains.

\textbf{Regularizing program length.} 
Unsurprisingly, overparametrizing by increasing the max program size  generates excessively long programs.
Because length is a proxy for complexity, these programs might also tend to be more overfit, and also harder for humans to understand and interpret.

To combat the code explosion caused by overparametrization,  we incorporate an additional term in our loss which penalizes the average program length.
Calculating program length from our parametrization of $\alpha$ is straightforward to do in linear time using dynamic programming, and is also a smooth, differentiable function of $\alpha$'s components.
Hence we can simply add the length-penalizing term to our loss.

\begin{figure}
\includegraphics[width = \columnwidth]{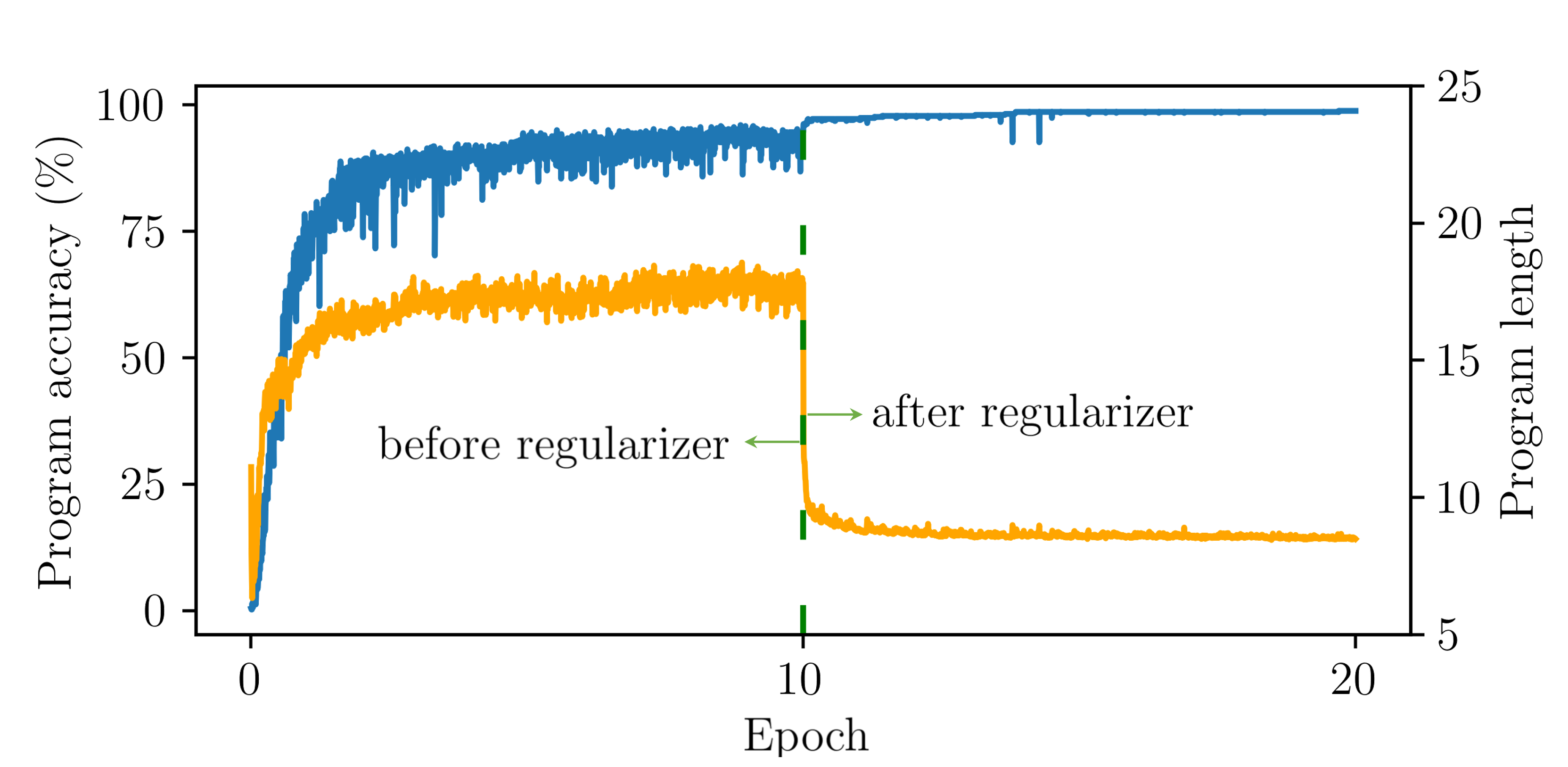}
\includegraphics[width = \columnwidth]{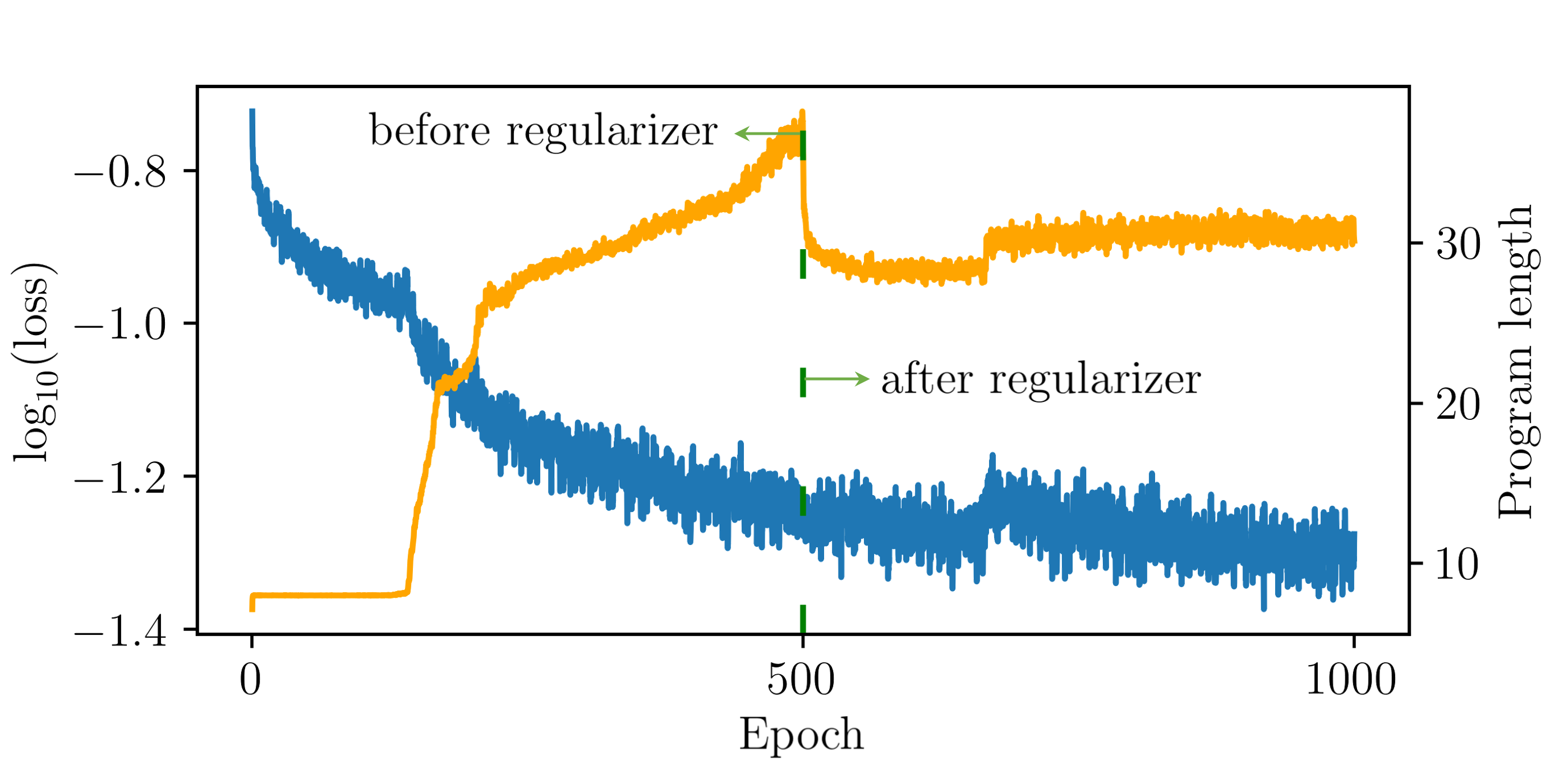}
\caption{Regularizing program length halfway through training refactors the programs to be shorter (orange) without sacrificing accuracy (blue). Top: CIFAR-MATH. Bottom: 3D graphics.}\label{fig:regularization}
\end{figure}

In practice we train ROAP without regularization for the first half of its training process--to encourage exploration--and then turn on this regularizer halfway through to compress and optimize the programs (Fig.~\ref{fig:regularization}).

\textbf{Min-sampling.}
The policy acts as a search heuristic that stochastically proposes programs.
Running the policy multiple times per task and taking the sampled program with the minimum loss allows trading more compute for lower loss, a trick we call `Min-sampling'.
This is conceptually related to importance  reweighting of samples from neural recognition models~\cite{burda2015importance}.

\section{Experiments}

\subsection{CIFAR-MATH} 
The classic warmup problem for neurosymbolic systems is to train an MNIST classifier by supervising only on the result of running an algorithm on that classifier's outputs.
For example, DeepProbLog~\cite{manhaeve2018deepproblog} and Scallop~\cite{NEURIPS2021_d367eef1} both train a digit classifier given examples of handwritten digits being added together:
Given examples like \img{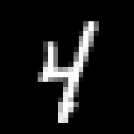}+\img{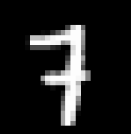}$\to $11, together with the logic of addition, these systems reason backward through the addition operator to train a neural network MNIST classifier.

CIFAR-MATH makes this warmup domain harder along several dimensions.
First, we introduce program synthesis  by not telling the system what arithmetic expression is executing on the input images: this system must infer and synthesize the correct symbolic equation.
Second, we consider more complex equations with several arithmetic operators.
Last, we switch from MNIST to CIFAR-10, and do  not tell the system that there are only 10 digits.
Thus the system has a harder reasoning challenge, because it has to reason backward through more complex expressions; a new induction challenge, because the expressions are hidden; and a nontrivial perception challenge, because 
CIFAR-10 is more visually complex than MNIST.

Following Fig.~\ref{fig:setting},  each task has a different \emph{hidden equation} (Fig.~\ref{fig:setting} illustrates $x+yz$).
The inputs to the equation are presented as CIFAR-10 images.
Each of the ten CIFAR-10 categories (dog, boat, frog, ...) is mapped to a different digit from 0--9, but this mapping is never given to the system.
Architectures $\alpha$ are built from a Domain Specific Language containing addition, multiplication, and subtraction.
The shared continuous parameters $\theta$ are the weights of a CNN that maps a CIFAR-10 image to a scalar.

We are interested in a variety of research questions, and compute evaluation metrics to help us answer each of them:
\begin{itemize}
    \item Can we synthesize the correct program? Because CIFAR-MATH comes with ground truth hidden programs, we check if the synthesized programs generate the same outputs on random inputs.
    \item Do we successfully learn a neural perception module, equivalently did the CNN learn a  CIFAR-10 classifier as a side effect of the overall training procedure?
To evaluate this we snap the CNN outputs to the nearest integer and report how often this yields the correct integer.
For example, if frogs correspond to the number 3, then correctly classifying a frog means predicting a number in the range $[2.5,3.5)$.
    \item  To understand if the learned programs generalize out of sample,  we designate one task to be trained only on small numbers (0--5), and then check if the synthesized neurosymbolic program extrapolates to larger numbers (6--9; multitasking makes it see these numbers on other tasks). 
    
\end{itemize}
\begin{table*}[h]
    \centering
    \caption{Experimental Results on CIFAR-MATH. min-sampling: 3 samples/gradient step}
    \label{tab:cifar_math.perf}
    \setlength\tabcolsep{3pt} 
    \begin{tabular}{cccccccc}
    \toprule 
         & Program-Acc & Symbolic-Acc & Test-Symbolic-Acc & Loss & Test-Loss & OOD-Loss \\
    \midrule
        ROAP (ours) & 91.8\% & 82.6\% & 48.7\% & 0.005 & \textbf{0.11} & 0.07 \\
         + min-sampling & \textbf{99.8\%} & \textbf{100\%} & \textbf{69.4\%} & 2.4e-4 & \textbf{0.11} & \textbf{0.05} \\
         + min-sampling; contiguous-image & \textbf{99.2\%} & \textbf{100\%} & \textbf{72.2\%} & 7.3e-4 & \textbf{0.12} & \textbf{0.04} \vspace{-2pt}\\ 
    \midrule 
        w./o. program & 0.0\% & 4.2\% & 4.7\% & \textbf{1.3e-5} & \textbf{0.11} & 0.52 \\
        w./o. amortized inference & 28.4\% & 49.5\% & 27.9\% & 0.022 & 0.16 & 0.15 \\ 
        w./o. gumbel-softmax & 21.0\% & 1.8\% & 0.9\% & 0.007 & 0.14 & 2.07 \\ 
        w./ Syntax-Tree parametrization & 69.0\% & 43.0\% & 28.1\% & 0.013 & 0.14 & 0.22 \\ 
        w./ max lines=10 & 11.6\% & 9.9\% & 6.9\% & 0.027 & 0.14 & 2.12 \\ 
    \midrule 
        REINFORCE & 0.2\% & 0.0\% & 0.0\% & 0.59 & 0.65 & 0.64 \\ 
        Terpret & 0.9\% & 4.0\% & 3.4\% & 6.4e19 & 5.9e19 & N/A \\
        Terpret + multitasking & 7.4\% & 11.1\% & 8.4\% & 7.6 & 12.8 & 0.74 \\
        NEAR & 0.8\% & 8.3\% & 6.9\% & 0.059 & 0.18 & 0.93\\ 
        dPads & 2.0\% & 9.6\% & 9.2\% & 0.36 & 0.36 & 0.96 \\
        HOUDINI & 17.4\% & 13.1\% & 9.4\% & 0.015 & 0.09 & 0.60 \\ 
    \bottomrule
    \end{tabular}
    \setlength\tabcolsep{6pt} 
\end{table*}
Tbl.~\ref{tab:cifar_math.perf} shows the metrics relating to the above questions for our system as well as ablations and baselines.
We use a \textbf{REINFORCE} baseline, which uses the score function estimator instead of Gumbel-Softmax;
\textbf{NEAR}~\cite{shah2020learning}, which uses A$^*$ to search the space of neurosymbolic programs, and does not perform multitasking or amortized inference;
\textbf{dPads}~\cite{cui2021differentiable}, which improves upon NEAR; \textbf{Terpret}~\cite{gaunt2016terpret}, which directly optimizes the parameters of $\alpha$,  and does not perform multitasking or amortized inference
; and \textbf{HOUDINI}~\cite{valkov2018houdini},  which solves tasks sequentially via enumeration while sharing neural network parameters across tasks. 

Overall, we find that the full model can jointly learn to ground its visual input into discrete symbols (numbers 0--9), and then transform those discrete symbols using symbolic equations that the system itself infers.
None of the baseline neurosymbolic synthesis methods meet that criteria.\footnote{We also tried Scallop~\cite{NEURIPS2021_d367eef1}, a leading neural logic programming system, but its differentiable-top-k-proofs inference method did not terminate on CIFAR-MATH problems.
We suspect this is because it has to build a massive proof tree when the arithmetic equation is unknown.}
We also find that a symbolic program aids out-of-sample generalization, as can be seen by comparing with the `w/o program' baseline, which replaces the program architecture with a small neural network.
This suggests ROAP has learned an appropriate division of labor between its CNN and its symbolic programs, with the CNN handling perception (but not reasoning) while the symbolic programs handle reasoning, thus enabling it to extrapolate out-of-distribution.

We additionally verified that ROAP does not  need the  image to be split  into three separate images showing each digit.
When the input is presented as a single contiguous image, our model's performance is essentially unchanged (Tbl.~\ref{tab:cifar_math.perf}, `contiguous-image').

\textbf{Why does the model learn the `right' latent symbols?}
A single CIFAR-MATH problem is ill-posed: it is not clear what latent symbols the neural network should output, because they are reprocessed by a (latent) program.
Our experiments establish however that the system readily converges on the `right' symbol grounding by mapping each CIFAR-10 image category to its corresponding digit.
Intuitively, this happens because multitasking introduces extra constraints on the function learned by the neural component, which has to serve a variety of downstream symbolic computations.

\begin{figure}[t]
    \centering
    \includegraphics[width=0.4\textwidth]{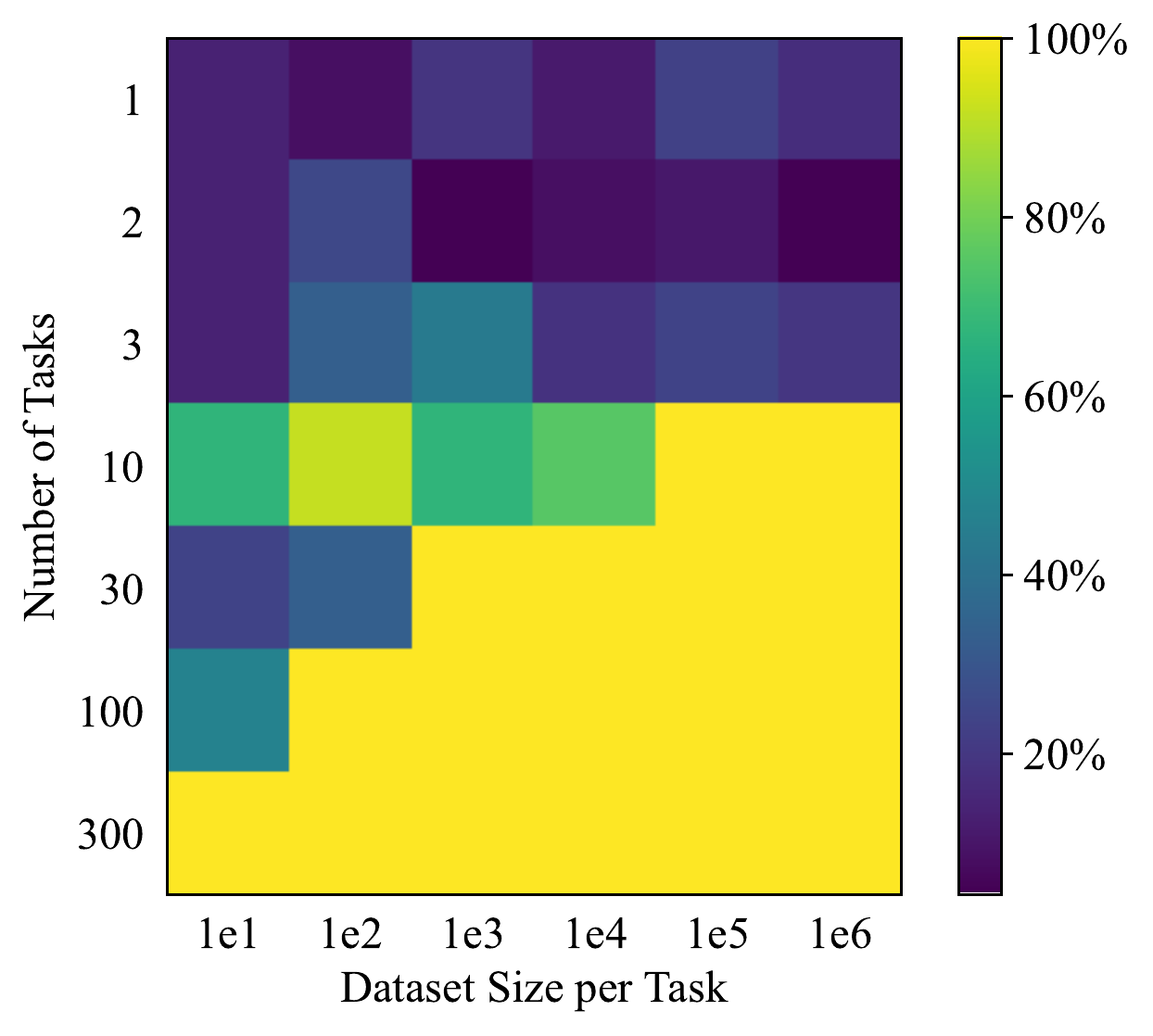}
    \caption{Effects of constraints on symbol grounding. Both having more tasks, and having more input-outputs per task, introduced added constraints. Heatmap shows max Symbolic-Acc over 5 runs, and should be approximately interpreted as whether or not it has a good chance of converging correctly in 5 runs.}
    \label{fig:cifar-math-heatmap}
\end{figure}

If this story is true, then the ability of the system to converge on a good symbol grounding hinges on having a sufficiently constrained optimization problem.
Extra tasks impose extra constraints, but so does having more input-output examples for each task.
We therefore study whether either of those constraints suffice for recovering the correct symbol grounding.
Fig.~\ref{fig:cifar-math-heatmap} shows success in recovering the correct symbolic basis as a function of the constraints imposed on the optimization problem, both by multitasking and input-outputs per task.
We see a phase-transition like structure where, once the total number of constraints passes a tipping point, the system `snaps' into the expected symbolic basis.
This shows the importance of constraints, and also that there is a tradeoff between the number of tasks, and the number of examples per task.

\begin{table*}[ht]
    \centering
    \caption{2D results. (NEAR's pathological behavior on these problems is to loop forever because it can fit a shape arbitrarily well with increasingly long partially completed programs, thus it never terminates with a completed program. HOUDINI's enumeration is inapplicable due to continuous parameters in affine transforms)}
    \label{tab:2dmetrics}
    \begin{tabular}{ccc|cc}
    \hline 
    \multicolumn{3}{c|}{Method} & \multicolumn{2}{c}{Chamfer Distance} \\ 
    \hline
    Name  & Train-mode & {Test-beam-size} & {No-refinement} & {Test-time-refinement} \\
    \hline
    CSG-Net & Supervised+RL & 10 & 1.14 & 0.41 \\ 
    CSG-NETSTACK & Supervised+RL & 10 & 1.02 & 0.34 \\ 
    PLAD & LEST+ST+WS & 10 & 0.811 & - \\
    UCSGNet & Unsupervised & 1 & 0.32 & - \\
    \hline 
    REINFORCE & RL & 1 & inf & - \\ 
    NEAR & Unsupervised & 1000 & N/A & \multicolumn{1}{c}{Pathological behavior} \\ 
    HOUDINI & Unsupervised & - & N/A & N/A \\ 
    Terpret & Unsupervised & 1 & N/A & 4.76$\pm$2.22 \\ 
    \hline 
    ROAP (ours) & Unsupervised & 1 & \textbf{0.21} & - \\
    \hline
    \end{tabular}%
\end{table*}

\subsection{Graphics Program Synthesis}

We use ROAP to synthesize neurosymbolic graphics programs.
We consider the problem of \emph{reconstruction}, which means inferring the shape of an object given a partial/occluded observation.
The graphics programs start with basic parts, like boxes and balls,  which are transformed and combined to generate 3D geometry.
What makes these graphics programs neurosymbolic is that, instead of hardcoding these basic parts, we allow the system to learn its own part library.
Each learned part is a simple shape that can be viewed as the output of a (tiny and unusual) neural network, whose parameters comprise $\theta$ (Appendix Sec.~\ref{sec:appendixreconstruction})

Our Domain Specific Language for graphics programs includes the ability to intersect and union shapes; reflect shapes over principal axes; a \texttt{for} loop that repeatedly translates its loop body; and affine transformations upon basic parts.
Each graphics program is a function from a point in space ($\R^3$) to a boolean  indicating whether that point is inside or outside of the object.

\begin{figure}[]
    \centering
    \includegraphics[width=0.5\textwidth]{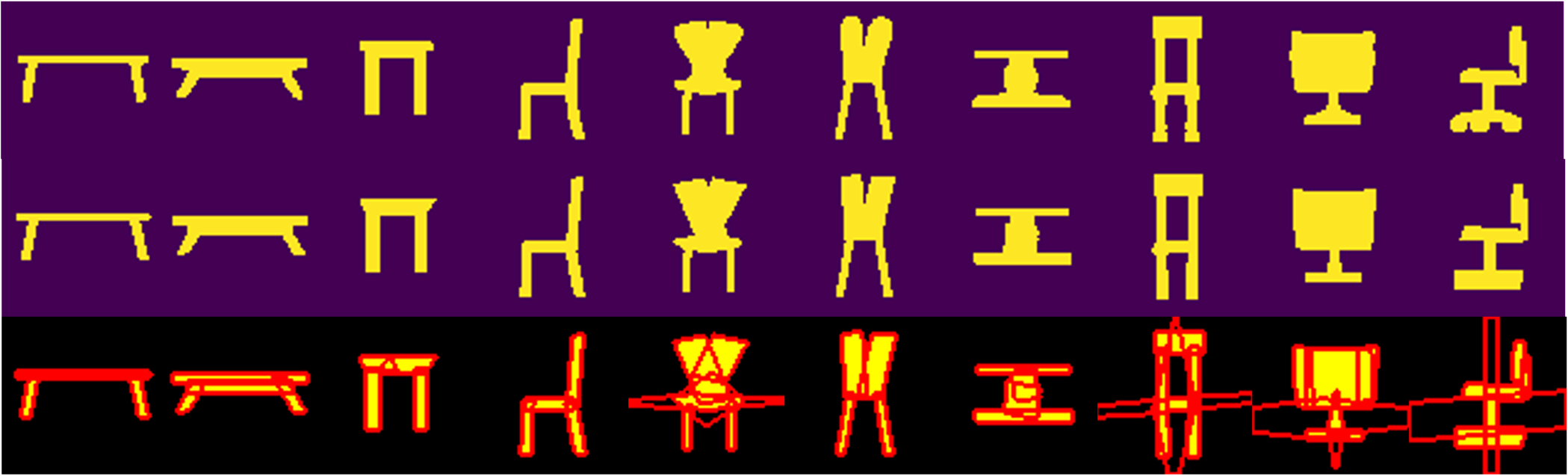}
    \caption{Qualitative results of 2D furniture silhouettes.
    Top, input silhouette. Middle, reconstruction. Bottom, parts used.}
    \label{fig:silhouettes}
\end{figure}

We first test on reconstructing 2D silhouettes of furniture (Fig.~\ref{fig:silhouettes}), which a series of recent graphics program synthesis works evaluate on~\cite{jones2022PLAD,PMID:33315554,kania2020ucsg,sharma2018csgnet}; see Appendix~\ref{sec:appendix2dexperiment}.
We assess reconstruction accuracy via Chamfer distance between the ground truth shape and the output of the synthesized program.
Tbl.~\ref{tab:2dmetrics} shows that ROAP achieves higher reconstruction accuracy compared to these comparable recent works.

\begin{table}[]
    \centering
    \caption{Experimental Results on 3D 
    }
    \begin{tabular}{ccc}
    \toprule 
    & Full & Crop-Plane\\
    \midrule 
        ROAP (ours) & 1.7 & 1.8 \\
    \midrule 
        w./o. program & 1.2 & 2.0 \\
        w./o. amortized inference & N/A & N/A \\ 
        w./o. gumbel-softmax & 9.5 & 8.4\\ 
        w./ Syntax-Tree & 2.0 & 7.1 \\ 
        w./ depth=10 & 8.7 & 2.7 \\ 
        w./ depth=3 & 2.7 & 13.1 \\
    \bottomrule
        
    \end{tabular}
    \label{tab:3dmetrics}
\end{table}

\begin{figure}[]
    \centering
    \includegraphics[width=\columnwidth]{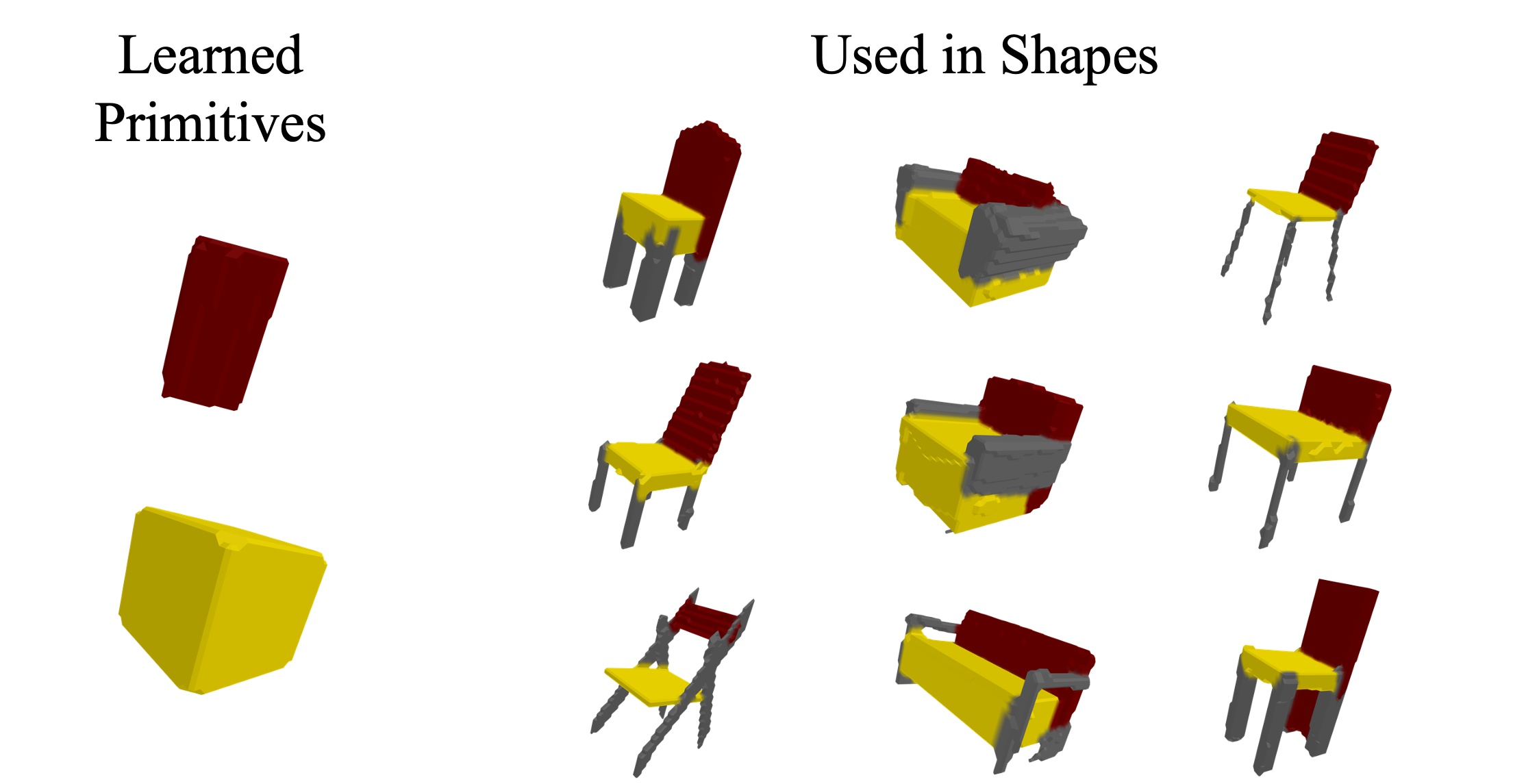}
    \caption{Neural part learning for 3D}
    \label{fig:3dparts}
\end{figure}

\begin{figure*}[b]
    \centering
    \includegraphics[width=\textwidth]{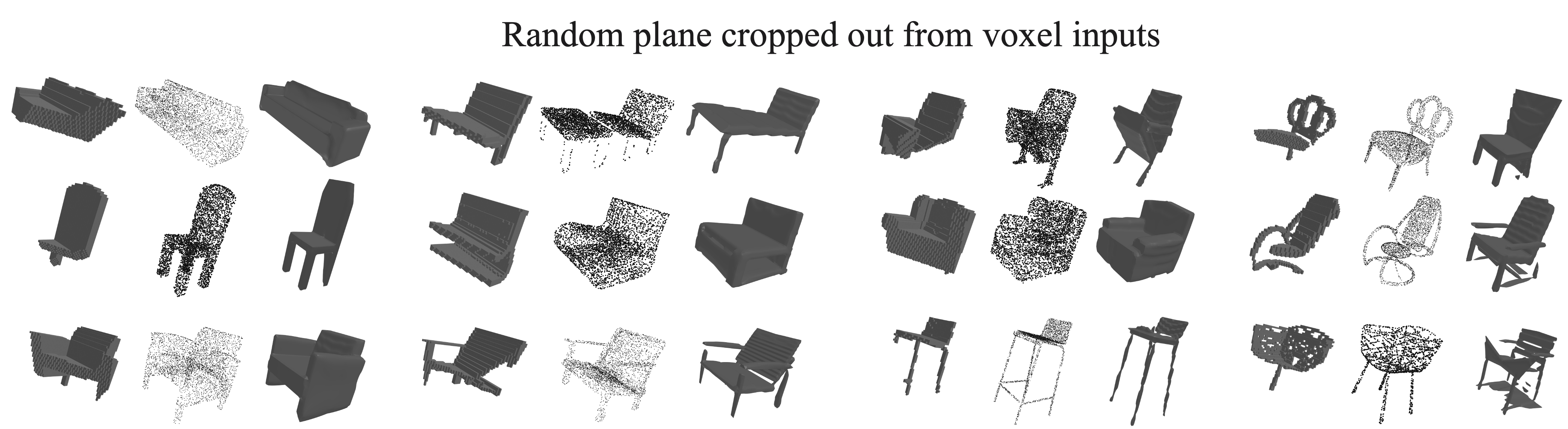}\\
    \includegraphics[width=\textwidth]{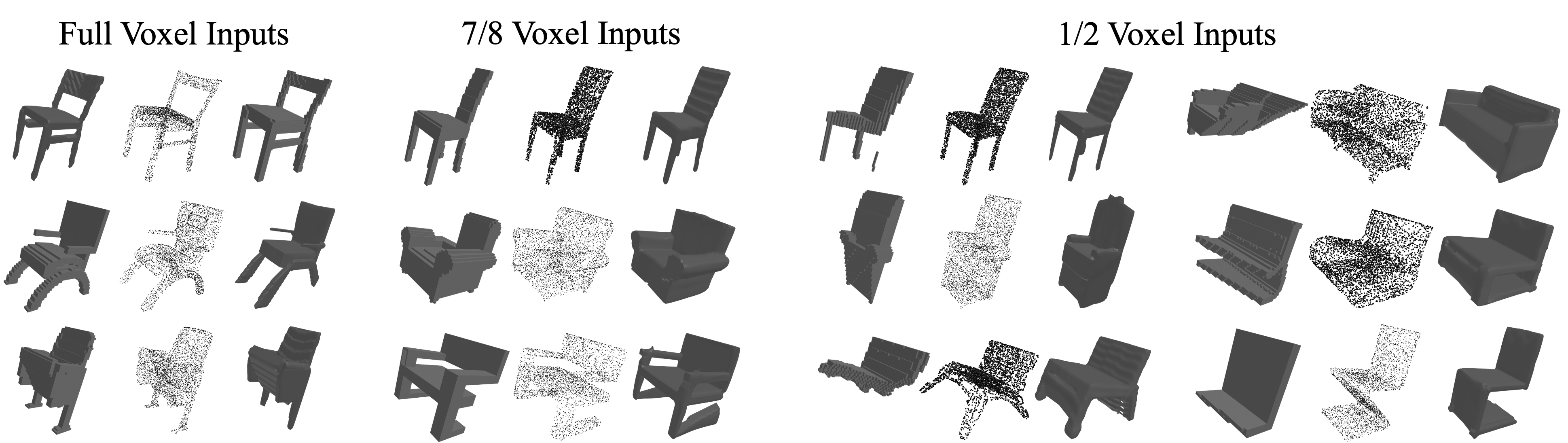}
    \caption{Results on 3D where the system inputs a voxel field with a random subset 
 cropped out (left models), from which it synthesizes a program (right models) that approximates the ground-truth shape (middle point cloud).
 Our model can be trained to complete partial geometry with a variety of different subsets taken out.
 Cropping out more of the input voxels makes the problem harder.
 Tbl.~\ref{tab:3dmetrics} reports quantitative results contrasting the easiest regime (full voxel inputs) vs the hardest regime (an entire random half plane cropped).}
    \label{fig:3d-plots}
\end{figure*}

Next we evaluate on 3D models.
Because ROAP does not supervise on ground-truth programs, we can apply it to datasets not designed for program synthesizers, and so we choose the canonical ShapeNet dataset~\cite{shapenet2015}; see Appendix~\ref{sec:appendix3dexperiment}.
Our goal is to study the qualitative behavior of our system and contrast with other general-purpose program synthesizers, \emph{not} to set a new state-of-the-art for ShapeNet.
ShapeNet has received over 7 years of attention from the deep learning and computer vision communities, who have built sophisticated yet specialized 3D reconstruction networks and training regimes (\citet{genova2020local} is representative in its sophistication).

Each reconstruction task is specified by a partial observation of a voxel field, and the synthesized program describes how to generate the underlying complete 3D shape.
We consider  corrupting the input voxel field by cropping out a random half-plane (Fig.~\ref{fig:3d-plots}).
Fig.~\ref{fig:setting} diagrams how ROAP represents shapes as programs that algebraically combine basic parts.
Fig.~\ref{fig:3dparts} illustrates example learned parts: Rather than preprogram boxes, cylinders, etc., the system learns from the data which primitives are most suitable.
It learned, for example, a boxy cuboid with rounded corners for modeling chair/sofa seats (yellow), and an elongated cuboid with a subtly curved top for modeling backs and headrests (red).
This data-driven discovery of basic symbolic abstractions was done without supervising on programs or part decompositions.

Last, we again quantify reconstruction quality via Chamfer distance, and compare against ablations of our system (Tbl.~\ref{tab:3dmetrics}).
The most important ablation of our system is `w/o program'.
This replaces the program with a neural network, essentially modeling an occupancy network~\cite{Occupancy_Networks}, which is a foundational  deep learning architecture for 3D reconstruction.
Given a complete observation of the shape, a pure neural network is superior at reconstruction (`Full' in Tbl.~\ref{tab:3dmetrics}).
But given a partial observation, the neurosymbolic program comes out ahead (`Crop-Plane' in Tbl.~\ref{tab:3dmetrics}).
This is because the symbolic program structure has an inductive bias primed to recognize symmetries and repeated parts.
Hence this high-level symbolic prior helps impute missing observations.

	\section{Related Work}
	Neurosymbolic programming is a growing area that seeks to engineer learning and inference methods for hybrid program/neural architectures~\cite{PGL-049}, and our work is a special case of this broad framework.
	Specifically, we tackle inductive program synthesis~\cite{gulwani2017program}--synthesizing programs from input-output examples--but where the inputs are continuous and must be preprocessed by neural networks into symbolic form.
	Prior works in this setting assume a hand-engineered inventory of basic symbols~\cite{ellis2017learning}, while others backpropagate through differentiable programs to jointly train network weights and program structure~\cite{gaunt2017differentiable}.
	Multitasking is a known strategy for this setting~\cite{valkov2018houdini}.
	Training a neural network to help guide search for discrete programs (amortized inference) is standard~\cite{shi2021crossbeam, chen2018execution}, and we extend that idea to continuous relaxations of program spaces. 
	
	The difficulties of gradient descent over relaxed program spaces is well known, to the extent that it has been dubbed the so-called `terpret problem'~\cite{gaunt2016terpret}.
	Unfortunately, such an approach is the most straightforward way of training neurosymbolic programs. From a technical perspective, our work hopes to make progress on the `terpret problem', thereby unlocking scalable and reliable training of this class of neurosymbolic programs.

 Some of our core tricks have debuted in prior neurosymbolic program synthesizers: Terpret noticed overparametrization helps~\cite{gaunt2016terpret}, Memoized Wake-Sleep deployed amortized inference~\cite{mws}, and HOUDINI shares neural network parameters across tasks~\cite{valkov2018houdini}.
Our technical contribution is providing the mathematical and algorithmic framework which allows these tricks,  and more,  to be combined into the same end-to-end learnable system.
For example, we showed that the reparametrization trick~\cite{jang2016categorical} made amortization compatible with relaxation and gradient-guided search.
	
	More fundamentally, our efforts connect to the body of work on the `symbol grounding problem'~\cite{harnad1990symbol}:
	How does a system learn to `ground' abstract symbols (e.g., numbers, parts) in terms of their high-dimensional perceptual counterparts (e.g., images of digits)?
	This problem is especially difficult absent strong supervision on the meaning of each abstract symbol~\cite{chang2020assessing}, and ROAP considers a distantly supervised setting.
	Prior works consider a variety of orthogonal techniques to address symbol grounding~\cite{topan2021techniques}, including scaffolding with natural language~\cite{andreas2016neural,mao2021grammar}.

\section{Conclusion}\label{sec:conclusion}
Our goal is to make progress on basic neurosymbolic problems: starting from perception, and absent symbol-level supervision, how can we discover basic symbolic abstractions together with the symbolic programs which manipulate them?
Although our experiments confirm that ROAP might be on the right track for solving these problems, our method has important limitations.
ROAP cannot operate without a reasonably-sized training corpus of programming tasks, although the fact that it does not need to supervise on source code helps address this limitation.
Fundamentally, ROAP assumes end-to-end gradient descent is the right approach, which means that program execution must be relaxed and differentiated.
It is not clear that differentiable program induction can handle sophisticated programming constructs, such as data structures and recursion~\cite{DBLP:conf/iclr/FeserBGT17}, at least in its current form.
Thus we especially hope ROAP helps spur more fundamental progress on differentiable program relaxation techniques.

\paragraph{Acknowledgements.} We gratefully acknowledge the support of a Research Scholar gift from Google.

\bibliography{example_paper}
\bibliographystyle{icml2023}

\newpage
\appendix
\onecolumn

\section{CIFAR-MATH Experimental Details}
\subsection{Experimental Setup}

\paragraph{Neural network.} We use an 18-layer ResNet backbone~\cite{he2016deep} as the image encoder with an MLP decoder, whose parameters collectively comprise $\theta$. 

\paragraph{Program denotation.} Fig.~\ref{denotation} specifies how we execute program architectures $\alpha$ in this domain using a simple dynamic program.

\paragraph{Dataset.} We generate 500 arithmetic tasks with 3 input variables, containing up to 3 operators.
For each  arithmetic task we have 1e6 I/O pairs for each task for training and 1000 I/O pairs for each task for testing. 

\paragraph{Training.} 	We train models using the Adam~\cite{kingma2014adam} optimizer with a learning rate equal to 3e-4 and $\epsilon=$\text{1e-5} for 20 epochs. The program length regularizer is not applied until halfway through training with a coefficient of $\lambda=$1e-4, which is multiplied into the program length before it is added to the rest of the loss. The temperature for gumbel softmax is set to 1 in the beginning and changed to 3 from epoch 15 to minimize the error gap from the continuous approximations of programs near the end of training.

\begin{figure}[h]
		\begin{align*}
			\hspace{2cm} \denotation{\alpha}_\theta(x)&=\text{Exec}_\theta(\alpha, x, L+V)\text{\phantom{test}\emph{execute program and extract output on line }}L+V\\
			\text{Exec}_\theta(\alpha, x, l)&=\text{CNN}_\theta(x_l)\text{, whenever }l\leq V\text{\phantom{test}\emph{load variables as first lines of code. We have V variables}}\\
			\text{Exec}_\theta(\alpha, x, l)&=\sum_o \alpha^O_{lo}\times F_o\left( x, \quad\sum_{1\leq a<l }\alpha^L_{la}\times\text{Exec}_\theta(\alpha, x, a), \quad\sum_{1\leq b<l }\alpha^R_{lb}\times\text{Exec}_\theta(\alpha, x, b) \right)\text{, whenever }l> V\\ 
			&\text{where }\alpha\text{ is a tuple of }(\alpha^O, \alpha^L, \alpha^R)\\
			F_1(x, A, B)&=A+B\text{\phantom{test}\emph{add}}\\
			F_2(x, A, B)&=A-B\text{\phantom{test}\emph{subtraction}}\\
			F_3(x, A, B)&=A\times B\text{\phantom{test}\emph{multiplication}}\\
			F_4(x, A, B)&=A\text{\phantom{testttetst}\emph{no-op/skip connection}}
		\end{align*}
		\caption{\textbf{Differentiable execution model} for a program sketch containing $L$ lines of code. $\alpha$ parametrizes the program via a triple of 2-dimensional arrays $(\alpha^O, \alpha^L, \alpha^R)$ containing values from 0-1. If $\alpha^O_{lo}=1$, then line $l$ of the program computes its value by executing operator $o$.
			If $\alpha^L_{la}=1$, then line $l$ of the program gets its left argument for the operator from line $a$.
			If $\alpha^R_{lb}=1$, then line $l$ of the program gets its rights argument for the operator from line $b$.
			The first $V$ lines of the program evaluate to input variables, and we assume that there are $V$ such variables and $L$ lines of code that follow. }\label{denotation}
	\end{figure}

\section{2D Reconstruction Experimental Details}
\subsection{Methods}
\subsubsection{CSG, Flip-Union, and For-Loop Operations}
\label{sec:2d-operations}

We now specify the denotation of graphics programs.
Recall that every graphics program is a function that takes a point in space ($\R^3$) to 0/1 depending on if that point is outside or inside the object.
In the relaxed semantics, we think of the denotation as producing a number in the range $[0,1]$.
We refer to such numbers in $[0,1]$ as `occupancy values'.

In general, the denotations of the graphics operations follows straightforwardly from their mathematical definitions.
For example, the union operator is represented as the maximum of its argument's denotations, i.e., $o=\max(o_\text{left}, o_\text{right})$. Specifically, the occupancy function of any shape that is a union of two parts is defined as $\denotation{\bigcup(z_\text{left}, z_\text{right})}(\vec p)=\max(\denotation{z_\text{left}}(\vec p), \denotation{z_\text{right}}(\vec p))$. The intersection operator is represented as the minimum of the occupancy values, i.e., $o=\min(o_\text{left}, o_\text{right})$, and the difference operator is represented as $o=\max(o_\text{left}-o_\text{right}, 0)$.

The flip-union and for-loop operations are implemented as unions of multiple sub-components. The occupancy values of the sub-components are determined by applying transformations to the coordinates that align with their semantics. Specifically, the flip-union operation is defined as: 
\begin{eqnarray*}
    \denotation{\text{flip-union}_{\theta_f}(z)}(\vec p) & = & \max(\denotation{z}(\vec p), \denotation{z}(\text{flip}_{\theta_f}(\vec p))),
\end{eqnarray*}
where $\theta_f\in \mathbb{R}^3$ defines a line by $\theta_f^T[\vec p;1]=0$ in the 2D case. Flipping a point against this line, i.e., $\text{flip}_{\theta_f}(\vec p)$ can be implemented using a simple affine transformation. Let $\theta_f=[a,b,c]$, then 
\begin{equation*}
    \text{flip}(\vec p) = \frac{1}{a^2+b^2}\begin{bmatrix}
b^2-a^2 & -2ab & 2ac\\
-2ab & a^2-b^2 & 2bc
\end{bmatrix}\vec p.
\end{equation*}

The for-loop operation is defined as 
\begin{eqnarray*}
    \denotation{\text{for-loop}_{\theta_{d}, C}(z)}(\vec p) & = & \max(\{\denotation{z}(\vec p-c\times \theta_d): c\in \{0,1,\cdots,C-1\}\}),
\end{eqnarray*}
which repeats the part $\denotation{z}$ for $C$ times by moving it in $\theta_d\in\mathbb{R}^2$ direction for $c$ times.

\subsubsection{Program Sketch}
The generation program is a union of three components: a simple component, a symmetry component, and a repeated component. The simple component, denoted as $\denotation{z_{\text{sim}}}$, comprises of various CSG operations. The symmetry component, $\denotation{\text{flip-union}_{\theta_f}(z_\text{sym})}$, is implemented using the flip-union operator. The repeated component, $\denotation{\text{for-loop}_{\theta_d,C}(z_\text{repeat})}$, is implemented using the for-loop operator. All sub-components $\denotation{z_{\text{sim}}}$, $\denotation{z_{\text{sym}}}$, and $\denotation{z_{\text{repeat}}}$, involve multiple CSG operations on simple primitive shapes such as squares and circles, using the straight-line-coding formulation. In addition, we incorporate gate parameters $\theta_{\alpha_\text{sym}}$ and $\theta_{\alpha_\text{repeat}}$ to control which components are included as 
\begin{eqnarray*}
    \denotation{\text{shape-program}}(\vec p) & = & \max(\denotation{z_{\text{sim}}}(\vec p), \alpha_\text{sym}\times \denotation{\text{flip-union}_{\theta_f}(z_\text{sym})}(\vec p), \alpha_\text{repeat}\times \denotation{\text{for-loop}_{\theta_d,C}(z_\text{repeat})}) \\ 
    \alpha_\text{sym} & = &  \textbf{1}(\theta_{\alpha_\text{sym}} >= 0) \\ 
    \alpha_\text{repeat} & = &  \textbf{1}(\theta_{\alpha_\text{repeat}} >= 0).
\end{eqnarray*}

\subsubsection{Balanced Training Loss}
To prevent models from becoming stuck in sub-optimal local optimums that fit only a portion of the training data, we employ a balanced training loss that adjusts the weights of samples based on the model's performance on them. Specifically, the loss is designed to as
\begin{eqnarray*}
    \mathcal L(\theta,\phi) & = & \mathbb E_{t\sim \mathcal T}\left[\text{weight}(t;\theta,\phi)\sum_{(x,y)\in \mathcal D_t}\text{Loss}(x,y;\theta,\phi)\right].
\end{eqnarray*}
We use the chamfer distance to measure the performances of models, $\text{cd}(t;\theta,\phi)$, and adjusts the weights of samples accordingly: $\text{weight}(t;\theta,\phi)=1-\max\left(1-\frac{\text{cd}(t;\theta,\phi)}{\text{threshold}}, 0\right)$. In practice, we set the threshold to be 0.95 based on our preliminary experimental results. 
\subsection{Experimental Setup}\label{sec:appendix2dexperiment}
We utilize the same CNN encoder and MLP decoder as UCSGNet~\cite{kania2020ucsg}. The program sketch includes one simple component, two symmetry components, and one repeated component. Each of their sub-components has 20 lines of codes in addition to the 32 transformed primitive shapes, including 16 circles and 16 squares. The maximum count of the for-loop operator is 3. 
We use the same dataset as UCSGNet~\cite{kania2020ucsg} which consists of 8000 CAD shapes in three categories, chair, desk, and lamps \cite{sharma2018csgnet}. 
\subsection{More Experimental Results}
\begin{figure}
    \centering
    \includegraphics[width=0.85\textwidth]{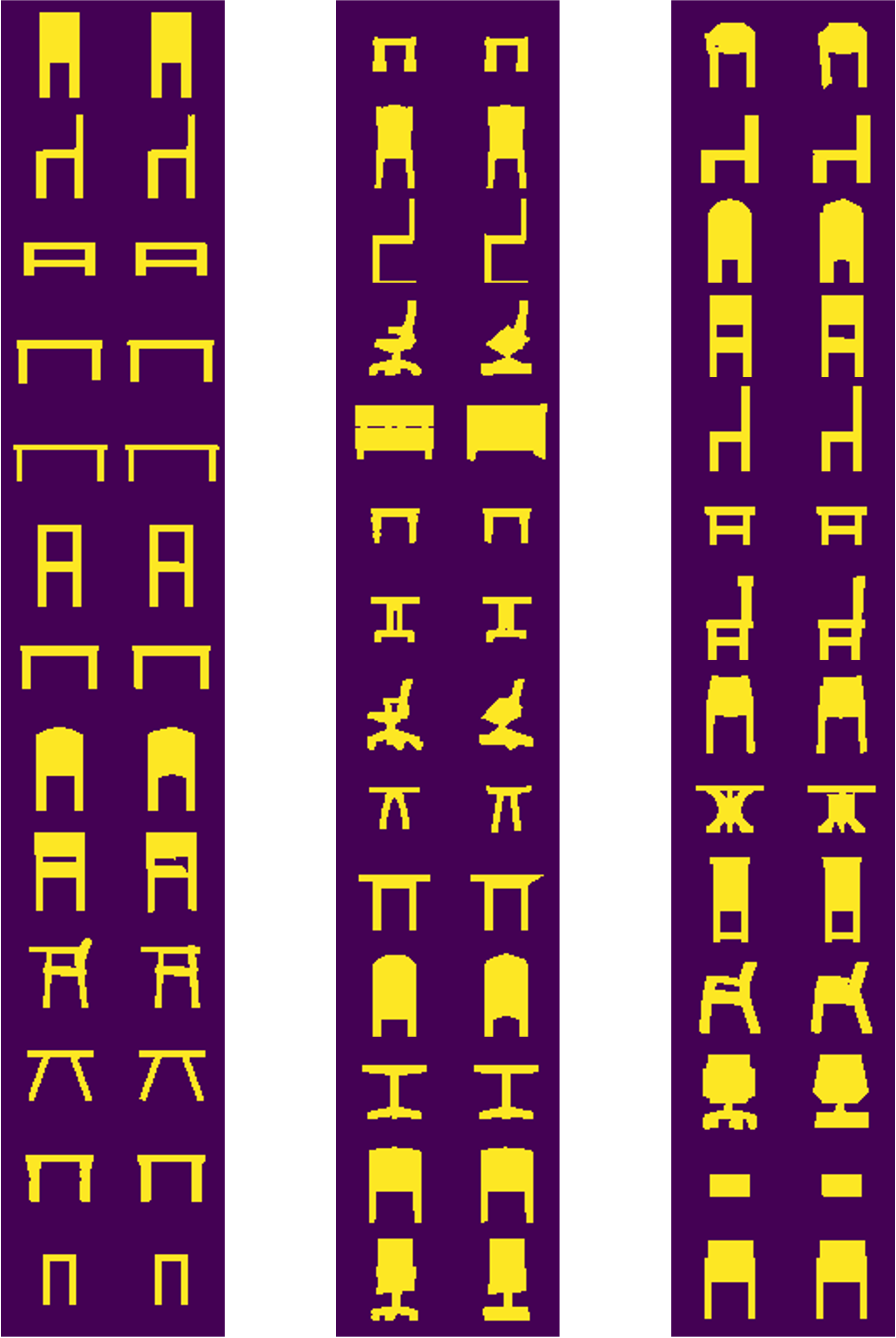}
    \caption{2D Results}
    \label{fig:app.2d}
\end{figure}

More reconstruction results are visualized in \cref{fig:app.2d}.

\section{3D Reconstruction}\label{sec:appendixreconstruction}
\subsection{Method}
\subsubsection{Quadratic Primitive Shapes}
Our program's expressive capacity is enhanced by the integration of curved primitive shapes, defined using a combination of two quadratic functions. The first function dictates the surface geometry, while the second performs a warp transformation. The signed distance function that defines the surface geometry is presented as follows:
\begin{eqnarray*}
    \mathcal D_{\theta_g}(\vec p) = \max( \theta_g^T[p_x^2, p_y^2, p_z^2, \abs{p_x}, \abs{p_y}, \abs{p_z}] - 1,~~ 2\abs{p_x}-1,~~ 2\abs{p_y}-1,~~ 2\abs{p_z}-1).
\end{eqnarray*}
This formulation ensures that the primitive shapes do not exceed the dimensions of a $1\times1\times1$ box by utilizing the last three linear surfaces. Additionally, the use of absolute value functions guarantees symmetry across the x, y, and z planes. This formulation not only allows for the representation of basic shapes such as boxes and spheres, but also enables the representation of more complex, curved primitive shapes with greater representation power.

Similar to affine transformations, quadratic warp transformations $f_{\theta_w}$ can be represented using a coordinate mapping function, as demonstrated below:
\begin{eqnarray*}
    f_{\theta_w}(\vec p)_x & = & \frac{p_x-t_x}{s_x} \\ 
    f_{\theta_w}(\vec p)_y & = & p_y \\ 
    f_{\theta_w}(\vec p)_z  & = & p_z, 
\end{eqnarray*}
where 
\begin{eqnarray*}
    \theta_w & = & [\theta_s; \theta_t; \theta_\alpha] \\
    \alpha_x & = & \textbf{1}(\theta_\alpha >= 0) \\ 
     s_x & = & \alpha_x \cdot (\theta_s^T[p_y, p_z, p_y^2, p_z^2, p_yp_z])+1\\
    t_x & = & \alpha_x \cdot (\theta_t^T[p_y, p_z, p_y^2, p_z^2, p_yp_z]).\\ 
\end{eqnarray*}
The warp transformation allows for the representation of irregular shapes, such as the mattock shape depicted in \cref{fig:setting}, by applying quadratic transformations to the coordinate $x$ based on the coordinates $y$ and $z$. The parameter $\alpha_x$ serves as a gate function that controls the degree of transformation. Note that, due to the symmetry properties of quadratic surfaces, the transformation of $x$ is equivalent to transforming $y$ and $z$.

\subsubsection{Shape Library}
To enhance the efficiency of primitive learning and grounding, we incorporate a shape library that is shared across tasks. The library comprises 128 warp transformations and 128 quadratic surface formulations, resulting in a total of 16,384 primitive shapes. The shape library retrieval mechanism is designed similarly to vector quantization. In particular, the shape library include parameters $\Theta_w\in \mathbb{R}^{128\times\abs{\theta_w}}$ for warp transformations and $\Theta_g\in\mathbb{R}^{128\times \abs{\theta_g}}$ for quadratic surfaces. For each query $\vec q\in\mathbb{R}^{\abs{\theta_g}+\abs{\theta_w}}$, it will return a primitive shape with the warp transformation parameter $\Theta_w\left[\argmax{\Theta_w \vec q\left[:\abs{\theta_w}\right]}\right]$ and the quadratic surface parameter $\Theta_g\left[\argmax{\Theta_g \vec q\left[-\abs{\theta_g}:\right]}\right]$. To enable training via gradient descent, we implement a probabilistic relaxation of the retrieval mechanism:
\begin{eqnarray*}
     \alpha_w & = & \text{(gumbel-)softmax}(\Theta_w\vec q[:\abs{\theta_w}]) \\ 
    \theta_w & = & \alpha_w^T\Theta_w\\ 
     \alpha_g & = & \text{(gumbel-)softmax}(\Theta_p\vec q[-\abs{\theta_g}:]) \\ 
    \theta_g & = & \alpha_g^T\Theta_g.\\ 
\end{eqnarray*}
The implicit function of the softly retrieved primitive shape is then $\mathcal{D}_{\theta_g}(f^{-1}_{\theta_w}(\vec p))$.

\subsubsection{CSG and Flip-Union Operations}

The CSG operations are defined in the same way as in the 2D case as outlined in \cref{sec:2d-operations}. The flip-union operation is an extension of its 2D counterpart, with the added dimension of flipping points against a plane defined by $\theta_f[\vec p;1]=0$. In practice, we set $\theta_f[-1]=0$ to ensure that the plane passes through the origin.

\subsubsection{Program Sketch}
The shape generation program is a symmetry of a sub-component that involves multiple CSG operations on learned primitives retrieved from a shared shape library. The CSG operations are formulated as straight-line coding.
\subsection{Experimental Setup}\label{sec:appendix3dexperiment}
We utilize the same 3D-CNN encoder and MLP decoder as UCSGNet~\cite{kania2020ucsg}. The program sketch is a symmetry of a sub-component that has 20 lines of code in addition to the 128 learned primitive shapes, retrieved from the shared shape library. We use the preprocessed dataset ShapeNet provided by~\cite{chen2020bsp}. It includes $64^3$ volumes of voxelized shapes and samples  16384 points as a ground truth with a higher probability of sampling near the
surface for training. In the 7/8 voxelized input setting, we randomly crop one-octant of the voxels by setting their values to zero. In the 1/2 voxelized input setting, we randomly sample planes as $a,b,c\sim\text{uniform}(0,1)$ and setting half of the voxels to zero, i.e., those voxels with coordinates $\vec p$ such that $ap_x+bp_y+cp_z>0$.

\subsection{More Experimental Results}
\begin{figure}
    \centering
    \includegraphics[width=0.85\textwidth]{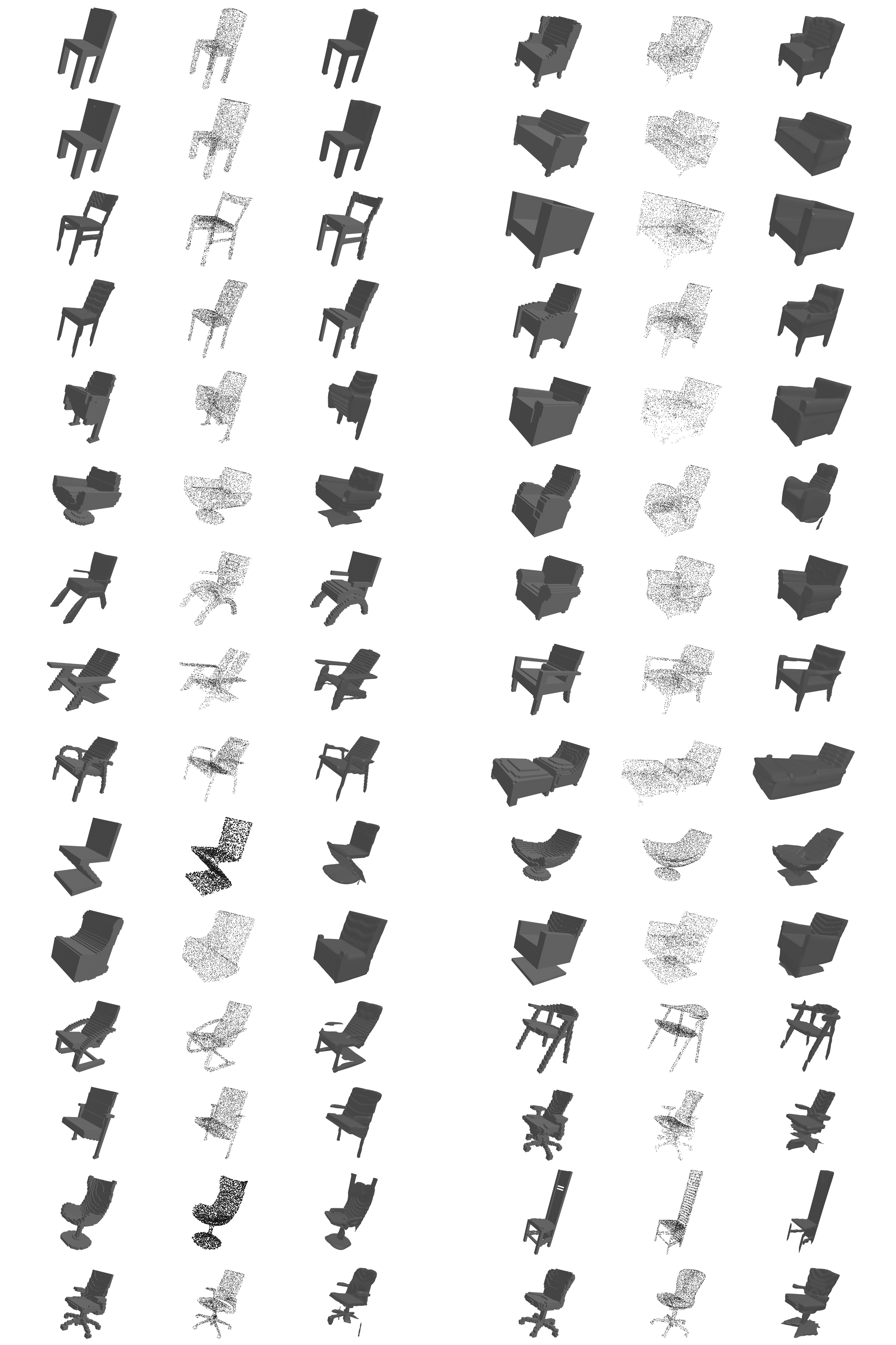}
    \caption{Full Voxelize Inputs}
    \label{fig:app.full-voxel-results}
\end{figure}

\begin{figure}
    \centering
    \includegraphics[width=0.85\textwidth]{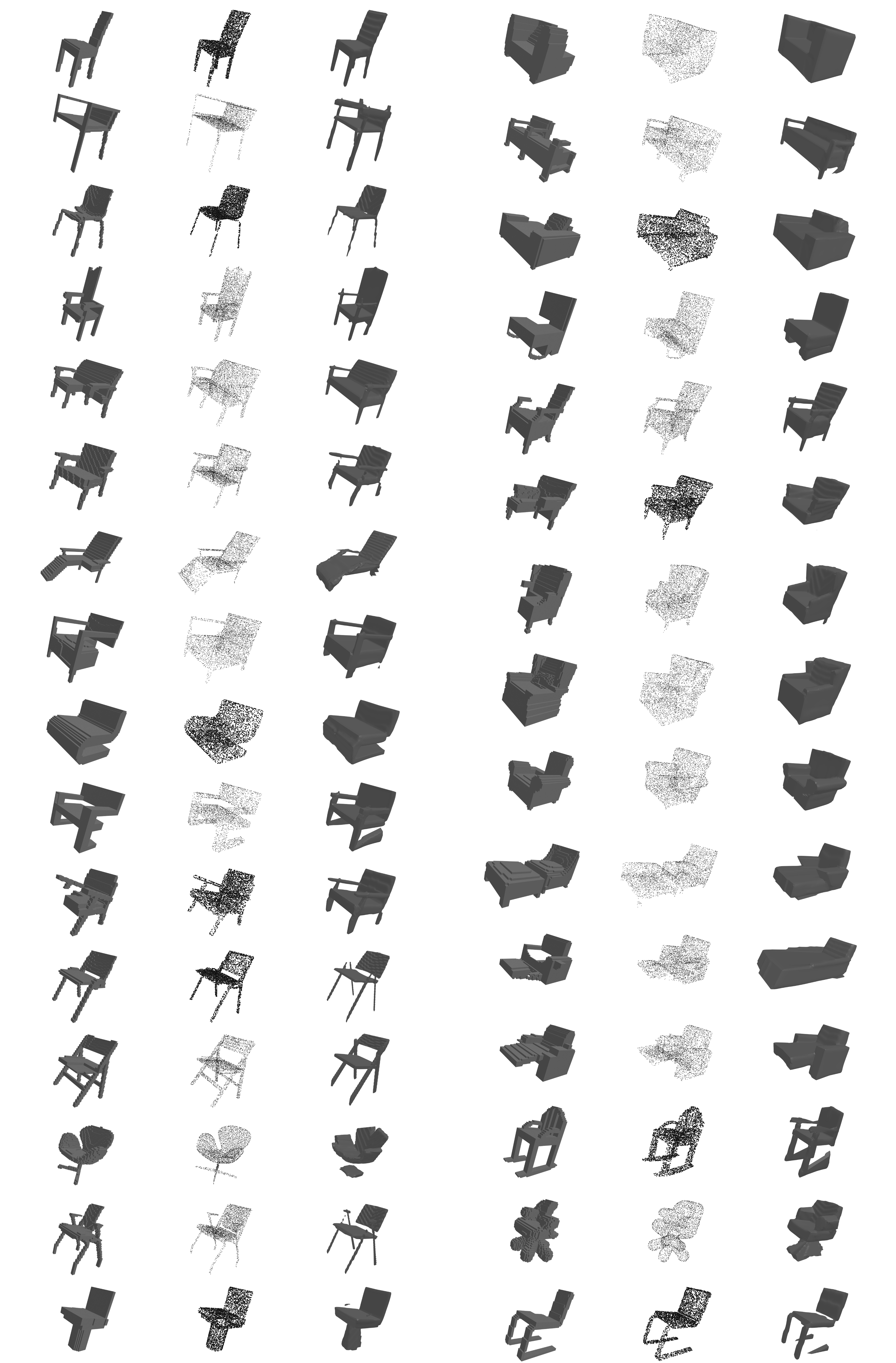}
    \caption{7/8 Voxelize Inputs}
    \label{fig:app.7/8-voxel-results}
\end{figure}

\begin{figure}
    \centering
    \includegraphics[width=0.85\textwidth]{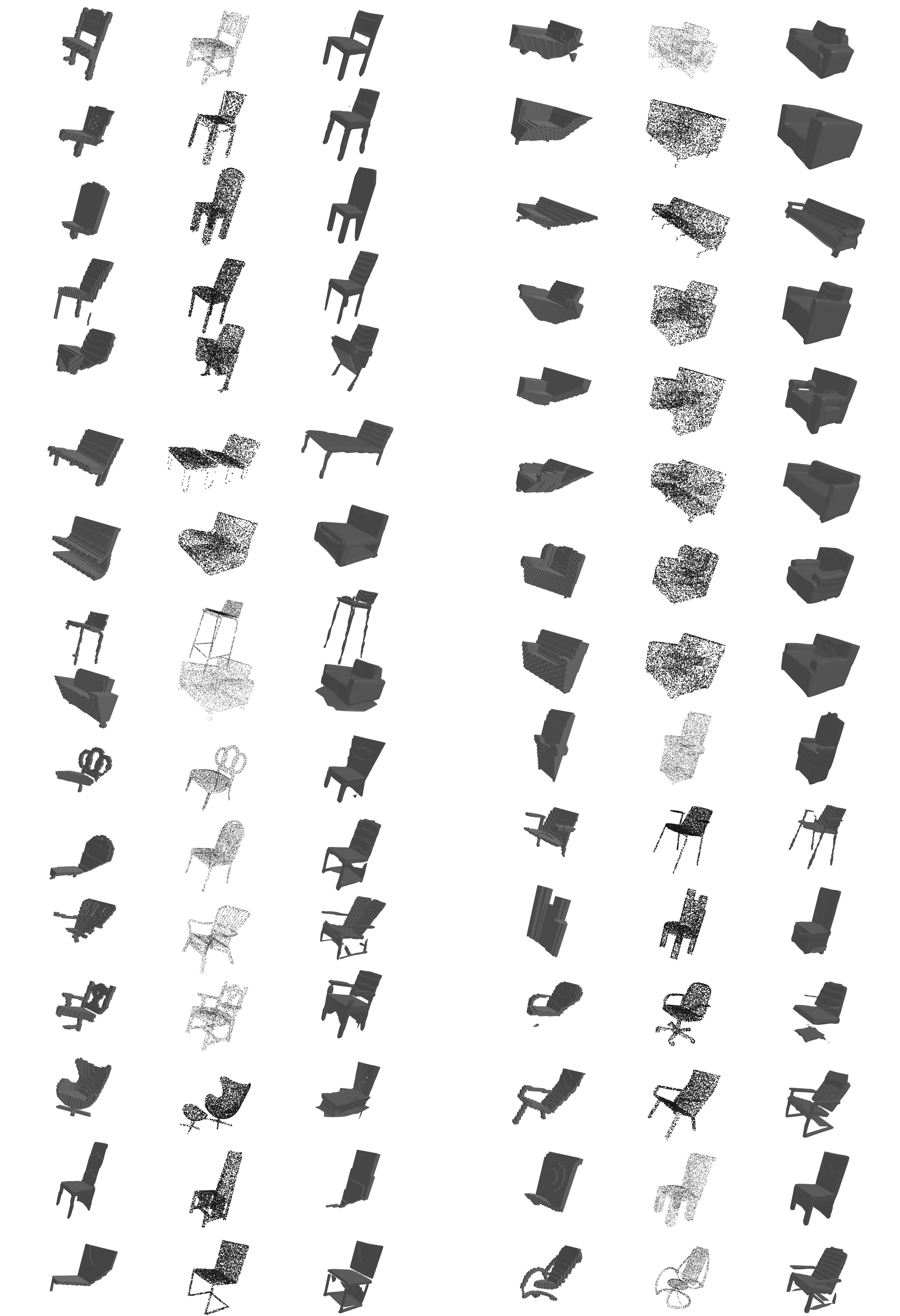}
    \caption{1/2 Voxelize Inputs}
    \label{fig:app.half-voxel-results}
\end{figure}

More reconstruction results are visualized in \cref{fig:app.full-voxel-results}, \cref{fig:app.7/8-voxel-results}, \cref{fig:app.half-voxel-results}.

\section{Theoretical Setting}\label{sec:appendixtheory}
We demonstrate the advantages of using over-parameterization in a simplified setting. Specifically, we consider a program synthesis algorithm that utilizes random search, and we assume that all programs are distinct and that operators take only one argument. Our goal is to find a program in the form of $p = o_L(o_{L-1}(\cdots(o_1(x))))$, where $L\ge 1$ is the length of the program, $o_l\in\mathcal O$ denotes operators, and $x$ is the input. There is no other program $p'\in\mathcal P$ such that $\forall x, p(x)==p'(x)$.

We represent the program sketch as a straight-line code with length $L'$, as shown in \cref{fig:parametrization}. Each line randomly selects one operator from the operator space $\mathcal O'$ and its argument from previous lines. We allow the program sketch to use the identity operator, resulting in $\mathcal O'=\mathcal O\bigcup \{\text{identity}\}$.

Due to the formulation of straight-line codes and the usage of the identity operator, there are multiple possible assignments of program parameters for the correct program. We calculate the exact probability of finding the correct program by randomly initializing the program assignment of the program sketch using dynamic programming. We set $\abs{\mathcal O}=3$ as the other experiments and show the probabilities of randomly initializing to a correct program for different correct program lengths $L$ and program sketch lengths $L'$ in \cref{fig:theory}.


\end{document}